# 关于机器翻译评测研究的综述性报告
# An Overview on Machine Translation Evaluation

**Lifeng Han**

data: February 22nd, 2022

**Abstract** 自从 1950 年代开始，机器翻译成为人工智能研究发展的重要任务之一，经历了几个不同时期和阶段性发展，包括基于规则的方法、统计的方法、和最近提出的基于神经网络的学习方法。伴随这几个阶段性飞跃的是机器翻译的评测研究与发展，尤其是评测方法在统计翻译和神经翻译研究上所扮演的重要角色。机器翻译的评测任务不仅仅在于评价机器翻译质量，还在于及时的反馈给机器翻译研究人员机器翻译本身存在的问题，如何去改进以及如何去优化。在一些实际的应用领域，比如在没有参考译文的情况下，机器翻译的质量估计更是起到重要的指示作用来揭示自动翻译目标译文的可信度。这份报告主要包括一下内容：机器翻译评测的简史、研究方法分类、以及前沿的进展，这其中包括人工评测、自动评测、和评测方法的评测（元评测）。人工评测和自动评测包含基于参考译文的和不需参考译文参与的；自动评测方法包括传统字符串匹配、应用句法和语义的模型、以及深度学习模型；评测方法的评测包含估计人工评测的可信度、自动评测的可信度、和测试集的可信度等。前沿的评测方法进展包括基于任务的评测、基于大数据预训练的模型、以及应用蒸馏技术的轻便优化模型。

**English Abstract**: Since the 1950s, machine translation has become one of the important tasks of artificial intelligence research and development, and has experienced several different periods and stages of development, including rule-based methods, statistical methods, and recently proposed neural network-based learning methods. Accompanying these staged leaps is the evaluation research and development of machine translation, especially the important role of evaluation methods in statistical translation and neural translation research. The evaluation task of machine translation is not only to evaluate the quality of machine translation, but also to give timely feedback to machine translation researchers on the problems existing in machine translation itself, how to improve and how to optimise. In some practical application fields, such as in the absence of reference translations, the quality estimation of machine

LH: PhD candidate (博士研究生，博士论文答辩与学位审核已通过，都柏林城市大学，爱尔兰)
ADAPT Research Centre, Dublin City University, Dublin 9, Ireland
E-mail: lifeng.han@adaptcentre.ie & lifeng.han3@mail.dcu.ie

translation plays an important role as an indicator to reveal the credibility of automatically translated target languages. This report mainly includes the following contents: a brief history of machine translation evaluation (MTE), the classification of research methods on MTE, and the the cutting-edge progress, including human evaluation, automatic evaluation, and evaluation of evaluation methods (meta-evaluation). Manual evaluation and automatic evaluation include reference-translation based and reference-translation independent participation; automatic evaluation methods include traditional n-gram string matching, models applying syntax and semantics, and deep learning models; evaluation of evaluation methods includes estimating the credibility of human evaluations, the reliability of the automatic evaluation, the reliability of the test set, etc. Advances in cutting-edge evaluation methods include task-based evaluation, using pre-trained language models based on big data, and lightweight optimisation models using distillation techniques.

**English Keywords**: Machine Translation; Machine Translation Evaluation; Human Evaluation; Automatic Evaluation; Meta-Evaluation

**1 简介**

机器翻译 (machine translation) 的研究始于 1950 年代 [152]，隶属于机器智能框架下的计算语言学 (computational linguistics) 的一个重要分支。机器翻译经历了基于规则理论模型 (rule-based)、基于实例的方法 (example-based)、基于概率统计学 (statistical MT, SMT)、和近年来的基于机器学习神经网络的方法 (neural MT, NMT) [18, 122, 32, 88, 33, 83, 151, 149, 91]。虽然机器翻译的质量持续改进，自动翻译的目标译文依然没有真正达到人类翻译专家的水平，这个现象在大部分语料对和不同领域的测试集上非常明显，最近的研究包括反应普遍流行的翻译测试集的狭隘性和文学领域 (literature domain) 机器翻译的表现很不佳 [95, 108, 77, 79]。因此，一如既往，机器翻译的评测 (MT evaluation, MTE) 扮演着推动机器翻译发展的重要角色 [77, 80]。机器翻译质量的评测本身是一个很有挑战性的研究课题，这源于翻译本身的多样性、语言的多变性和丰富性、以及语义相似度计算的复杂性。

这份报告包括对人工评测、自动评测、和针对评测的评测（元评测）的介绍、以及该领域一些前沿的研究进展，请参见图1，其中还包括交叉性的研究比如有人工参与的 Metric、以及 Metric 用于质量估计的研究。图1的上部分框架还揭示这个元评测的理论图也可应用于大部分的自然语言处理评测任务、不仅限于机器翻译。

有关机器翻译评测的国际赛事包括每年一届的统计机器翻译会议（WMT）[89, 21, 23, 24, 25, 26, 27, 12, 13, 14, 15, 16, 17, 8, 9, 10] 所组织的人工评测、自动评测（Metrics）和质量估计任务（QE），美国国家标椎和技术机构（NIST）组织的机器翻译比赛 [100][1]，和语音语言技术国际研讨会（IWSLT）[46, 124, 125, 49] 协办的文本翻译赛事；地区性的赛事包括中国机器翻译研讨

---
[1] https://www.nist.gov/artificial-intelligence

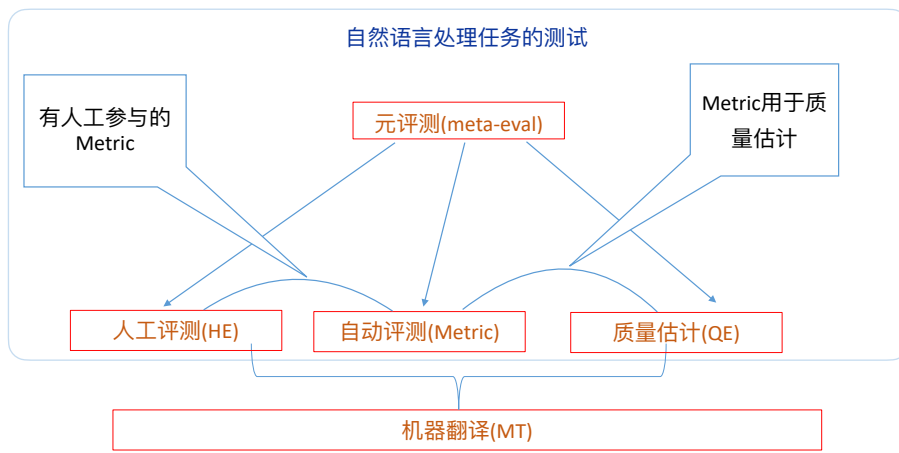

**Fig. 1** 机器翻译元评测概览

会（CWMT）。这份报告的大部分方法来自对以上国际和地区性的评测赛事的总结。

## 2 相关工作

从翻译教学和翻译工业应用的角度，[138] 在 2005 年做了有关机器翻译错误分类的研究。欧洲机器翻译研究联合项目 EuroMatrix 于 2007 年的一份报告简介了人工评测和当时流行的自动评测 [48]。美国国防先进研究项目机构（DARPA）的 GALE 项目助攻机器翻译并在 2009 年的一份汇报中介绍了自动评测和半自动评测，包含基于任务的和有人工参与的评测方法，其中 HTER 是该项目主要信赖评测指标。该报告还指出评测方法可用来机器翻译参数的优化 [43]。2013 年欧洲机器翻译会议（EAMT）的一份邀请报告阐述了该作者所在机构开发的 Asiya 在线机器翻译错误分析平台。同时还提及了机器翻译评测的简史，包含基于词面相似度的方法和语言学驱动的方法。这份报告区别于以上工作，在人工评测、自动评测、和元评测上分别加以综合介绍，并且对近几年的该领域研究进展进行更新讲解。此报告是基于我们近期发表在 "翻译建模：数字时代的翻译学 (MoTra21)" 国际研讨会的工作 [80]。

## 3 人工评测

人工评测部分我们分两个小节介绍传统的方法和后续发展的方法，参见图2。

### 3.1 传统方法

早期的机器翻译人工评价标准始于美国自动语言处理指导委员会 (ALPAC) [28] 所制定的 "清晰度" 和 "保真度"。清晰度被定义为：尽最大可能地，翻译

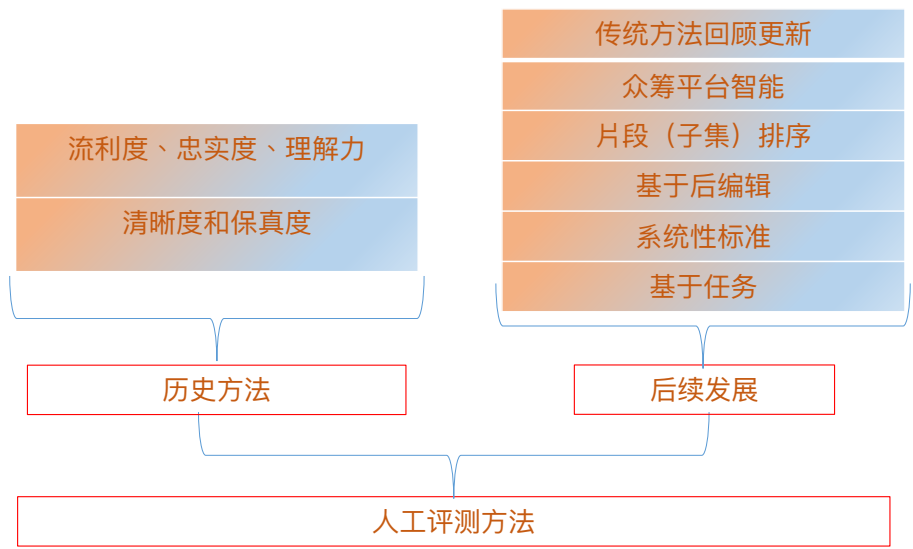

**Fig. 2** 机器翻译人工评测分类

文本应该读起来像正常的认真写出来的片段，并且容易理解，就像是一开始就是用目标语言所写的。保真度被定义为：翻译文本应该尽小可能地对源语言文本进行扭曲、歪曲、或者制造争议。

在 1990 年代，美国先进研究项目机构 (ARPA) 制定新的机器翻译评测标准，包含流利度、忠实度、和理解力 [34]。这些标准被后续机器翻译竞赛所采纳 [154]。流利度反应翻译文本的句法和语法正确性和流畅性，流利度的判断不需要参考原文；忠实度反应对原文的保真性，需要有源语言文本的指导；理解力反应信息度也就是看一个机器翻译系统能否输出给用户充分有效的和必要的信息。最初的流利度和忠实度的设计包含五个不同等级；而对于理解力，则设计了六个不同的问题让专业评判者回答。

由于流利度和忠实度的互补性和易用性，机器翻译研究人员对这两个指标进行了不同程度的应用、修改和整合等。比如以 "准确性" 作为整合的标准，[7] 对准确性加以分类，包括简单字符串、生成字符串、和解析树的准确性。[133] 的工作进行了流利度和所需字数的相关性计算来区分人工翻译和机器翻译。语言数据集团 (LDC)[2] 采用五个等级的流利度和忠实度来评估 NIST 的机器翻译比赛。其中对流利度的判断除了语法要求，还包含了对习惯用词（惯用语）的选择。

[144] 则对忠实度进行了四个等级的划分：非常、一般、较差、和完全不忠实。非常 (highly): 翻译文本非常信实的传达原文意思；一般 (fairly): 翻译文本在传达原文意思上一般表现一般，在字序、时态、语气、数字等方面有问题，或者存在重复、添加或遗漏字词；较差 (poorly): 译文没有足够反应到原文意思；完全不忠实 (completely not): 译文没有反应原文的任何意思。

---

[2] https://www.ldc.upenn.edu

3.2 后续发展

后续和近期发展的人工评测归为以下几类：基于任务 (task)、后编辑 (post-editing)、新标准、子集排序 (segment-ranking)、众筹平台 (crowd-sourcing)、和对传统方法的回顾更新。这种分法是为了便捷需要，有的人工评测方法可以涉及多个子类的交叉，比如基于任务和后编辑的两个子类型。

*3.2.1 基于任务*

于 1998 年 [153] 开发了一个基于应用任务的用于日文到英文翻译的针对性的评测方法，此研究倾向于参考美国国防先进研究项目机构（DARPA）的制定标准，设计依赖于语言对的措施，包括扫描、排序、话题识别、任务接受度、源语言模式类型、等技术。1999 年 [44] 提供了基于任务的机器翻译评测简介。2006 年 [150] 介绍了一份新的基于任务的机器翻译评测，通过抽取"主题（who）、时间（when）、和地点（where）"三个元素了评价其质量，这个工作并且应用于事件理解任务 [93].

*3.2.2 基于后编辑*

基于后编辑的方法思想是去比较从修改机器翻译的输出到达成正确翻译所需的努力，这个工作可以在两种不同情况下进行：1）有参考译文的情况下进行，去朝着参考译文来编辑翻译输出；2）也可以在没有参考译文的情况下进行，把机器翻译输出编辑成一个可以接受的正确的结果。第二种情况的完成会伴随着一个参考译文的生成。早期的这类工作包括 HTER，就是根据第二种设计，在没有参考译文的情况下，采用最优化的办法编辑自动翻译输出来计算所需的最小编辑量，编辑步骤包括 "曾词、减词、替换"[142]。这个方法的缺陷在于：虽然我们可以根据编辑量对不同机器翻译系统进行排序，但是我们并不知晓具体编辑量和翻译质量之间的准确联系。比如说、多少量的曾词、减词、和替换词代表什么样的翻译质量？有没有这样的统计规律？这还是一个未知的问题。

最近，我们自己提出的基于任务和后编辑的评测方法 HOPE[54] [3]，针对 HTER 的缺陷和现有一些其他复杂的人工评价比如 "多维度质量指标"MQM(multi-dimensional quality metric [106]) 进行改进，设计了八项目前机器翻译输出容易出现的错误类型，并对不同错误的程度进行区分，评测人员对翻译输出的错误类型按照参考分门别类并且给出错误程度分数，然后计算总分。目前设计的八个错误类型包括：影响度、调整度、术语、语法、错译、风格、矫正、名字。其中，影响度反应译文对原文的语义覆盖，调整度反应原文的部分内容需要适当调整或者修改（这源于原文出错或者应用的需要）但是机器翻译系统或者人工翻译人员没有做到而导致终极用户受影响。对每个翻译错误类型的程度的区分，HOPE 设计了三个不同等级，包含：零错误 (打零分)、微错（打 1-4 分）、和大错（打 5 分以上），这先在翻译片段（segment-level, 子集）上进行标注，片段可以是一个句子，或者一个段落，片段级别的错误分数惩罚系数在 HOPE 里面被记做 EPPTU（error point penalty of translation unit），它是片段级别的错误数量乘以相应各个程度系数的总和。然后系统级别的 HOPE 分数是所有片段级别分数的总和。

---

[3] 取名于 task-oriented human-centric metric based on professional post-editing

进一步的 HOPE 评测标准还推出了字词级别 (word-level) 的应用。也就是在 HOPE 分数计算的第一步，替换以句子为基本单位的"片段"级别为字词级别；在英文和其他西方语言这个可以是一个单词为单位，在中文和相近的亚洲语系这个可以是一个字为单位。在英语到俄语的两个不同数据集上（市场领域文本和商业调查文本）的初步试验证明 HOPE 评测方法简单易行，并且可以准确反应机器翻译的译文质量，区分不同机器翻译系统的表现。回应上一集提到的基于任务的模型（3.2.1），HOPE 也可以应用于评价"可以接受 (good enough) 的机器翻译"评价任务，比如在某些应用领域（用户指南、说明文件、旅游指导等），并不需要完美的机器翻译但要求达到一个可以接受的水平即可。HOPE 评价方法模型和所应用的语料数据已经置于在线开源状态，详见 https://github.com/lHan87/HOPE。

与 HOPE 比较相关的一个工作是本世纪开始时期于 2001 年美国自动化工程师协会（SAE）所开展的基于人工标注的机器翻译评价方法。SAE 制定了七项错误类型和两项子类型并且对不同类型进行了错误程度的区分。但是这个工作在机器翻译和评测领域并没有被广泛应用起来。这可能与其类型设计、可操作程度、和开源程度有关[4]。

HTER、HOPE 和 SAE 都可以归属于有人工参与的 Metric 类型，见图1。

### 3.2.3 子集（片段）排序

在每年的国际机器翻译研讨会所组织的评测比赛中，基于片段排序的方法被多年应用于人工对翻译质量的评价。在历史的比赛中，人工评价者被要求对五个机器翻译系统的输出从高到低的排序，这五个输出来自于相同的原文只是打乱了系统顺序从而让评价者无法知晓被评价的句子来自哪个系统 [26, 27, 12]。每次排序，原文、译文和参考文都会提供给评价者。对于每次排序，在没有平分的情况下会有 10 种成对比较的结果。成对比较的结果会转变成最终的系统排名分数。成对比较的结果可以用来判断一个系统赢得其他系统或者输给其他系统的频率是多少，参考一下公式：

$$\frac{\text{Wins}}{\text{All} - \text{Ties}} \quad (1)$$

其中 Wins 是赢的成对比较数，All 是总的成对比较数，Ties 是平局的成对比较数。

### 3.2.4 众筹平台智能

因为国际机器翻译研讨会 WMT 所汇报的人工评价的一致性分数在很多语言对上不佳，尤其是在片段子集级别上，很多研究人员开始探索新的更可靠的人工评测方法。其中被 WMT 所采纳的工作包括 [57, 58]，此工作分析认为 WMT 人工评价的一致性不高的原因可能是因为比赛组织者所设定的打分区间的问题，比如评价者可能对一个翻译结果的可选的两个毗邻的打分区间模

---

[4] SAE J2450 项目指出：对产品服务信息的低质量的翻译会导致很多风险，包括客户信任度、质量保证花费、对车辆的损坏、和对人的伤害等，因此一个非主观的连贯性的自动评测方法的应用可以更好的管理翻译质量控制。在 SAE 的评价设置里面，他们考虑到所花时间、费用、和人工投入等因素。见 https://www.sae.org/standardsdev/j2450p1.htm

棱两可，或者对两个机器翻译的好坏比较模棱两可。基于这个驱动，作者设计了连续性打分模式 CMS（continuous measurement scales），这个方法的实现是通过亚马逊 MTurk 众筹平台，召集网络上工作的人员，并且设置一些质量控制的技术，比如人为加入不好的参考译文、多次询问、以及统计学重要性区分等进行筛选高质量的打分人员和打分结果。更详细的质量控制方法在此工作的最新进展中有介绍 [59, 60].

在最近年的 WMT 比赛中，为了提高人工评价的质量和可信度，从众筹平台的人工评测开始向翻译或者语言学专家级别的人工评测转变，比如 WMT2020 年的质量估计比赛（QE2020[145]）。但是由于这个设定的花费较大，目前只是应用于有限的语言对。

*3.2.5 回顾更新传统指标*

2020 年 [128] 的工作批评传统模式的人工打分和系统排序的方法无法真正反应机器翻译的真正问题。此文提出让评价者标出译文的所有的翻译问题所在，无论是字词级别还是句子级别。给评价者的所需回答的问题多于译文的理解力（comprehensibility）和忠实度（adequacy）相关。在作者的后续工作里 [129]，通常出现的频率最高的错误类型会被总结下来。

*3.2.6 系统性标准*

在本节最后，我们转换评价的对象，除了对译文的准确性要求，针对机器翻译系统本身，[86] 于 2003 年提出了一些列系统性评价要求，其中包括适用性（suitability）：输出文本的准确性是否适用于具体测试的领域；互操作性：此机器翻译系统是否可以被其他软件或硬件平台使用；可靠性：是否会出现系统瘫痪现象以及需要多久恢复；使用性：是否有容易使用的平台、是否易学易用、看起来美观；有效性：对操作文本是否连贯处理；维护性：是否可以根据用户需要更新系统；转换性：随着资源和错误处理的更替，是否新的版本可以替代旧的版本。

**4 自动评测**

人工评测一直以来作为评测机器翻译质量的最终标准，但是人工评测也存在很多缺陷，比如耗时、昂贵、不可重复 (抑或不可重用) 性、以及很多情况下出现的人工评价人员之间的不一致性（主观性）。因此自动评测方法成为技术和实践上的双重需求。自动评价的产生伴随着几个不同的类型，包括需要参考译文的和不基于参考译文的情况。在需要参考译文的模型里又包括使用单个参考译文和多项参考译文的类型 [103, 66, 80]。基于参考译文的自动翻译评测模型，多属于计算自动译文输出和参考译文之间的相似度来评价翻译质量。当然，语言相似度的计算是一个很有争议、也很有挑战性的问题，比如句法上、语义上、风格上、写作领域和标准上的不同和变化等。不依赖于参考译文的评价模型大多依赖机器学习的特征模型，从源语言的原句字和目标语言的译文里提取有效特征来估计译文质量、这些特征可以包括词性、句法、语言模型等。与人工评价相比，自动评价的好处包括廉价、快速、可重复性、和可用来调整和优化机器翻译的模型参数等。

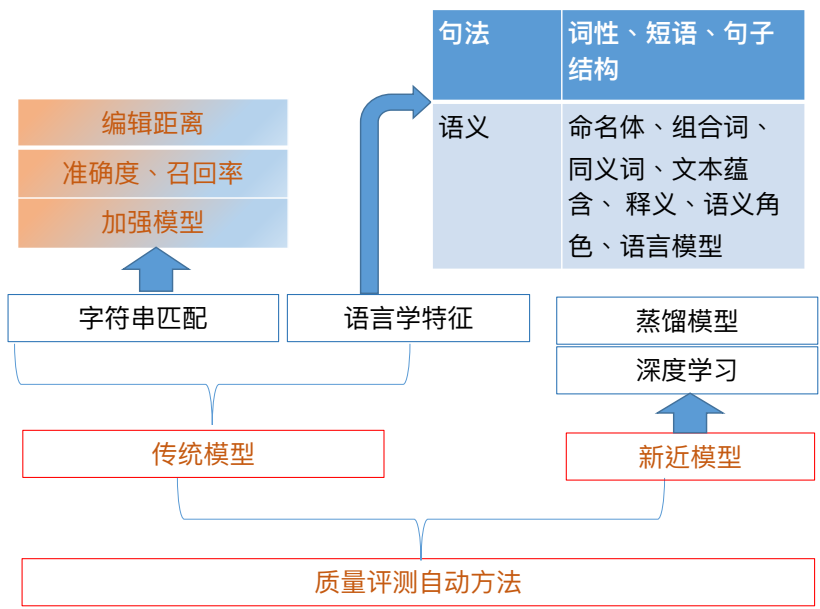

**Fig. 3** 翻译质量自动评价方法分类

在本节，我们将传统的自动翻译评价模型分为基于字符串匹配的（n-gram）和应用语言学特征的两类。在字符串匹配的种类里又包括基于编辑距离、准确度和召回率、以及加强模型的。在语言学特征上我们把基于句法和语义的分两个子类。其中句法特征包括词性、短语、句子结构等，而语义特征包括命名体、组合词、同义词、文本蕴含、释义、语义角色、和语言模型等。我们将基于深度学习和大规模预训练模型的评测方法归为新近模型一类。在这一个领域，最近又发展了优化大数据和大模型花费的蒸馏模型。总览见图3。当然这些分发是为了介绍和理解的方便，部分自动评测模型也会涉及到多个子类里面、各个子类也有时候会交叉，比如我们自己开发的 LEPOR、hLEPOR、和 nLEPOR 方法 [71, 74]，将会在下文提到。

### 4.1 字符串 (n-gram) 匹配

在本节我们介绍基于字符串 n 元语法 (n-gram) 的匹配计算，包含基于编辑距离、准确度和召回率、以及新近设计的次序计算因素的模型、和综合各类因子的加强模型。

*4.1.1* 编辑距离

1992 年 [146] 介绍了根据计算译文输出到参考译文的最小编辑距离来自动评价翻译质量好坏，起名单词错误率 WER(word error rate)。这里设计的编辑距离包括加词 (incertion)、删词 (deletion)、和替换词 (substitution)。WER 的设计是受到 1966 年所提出的莱文斯坦距离（Levenshtein Distance）[99] 公

式[5]。莱文斯坦距离的三个因素是加、删、和移动（位移），并且应用于任何字符串级别。WER 计算公式如下（总的编辑数除以参考译文长度）：

$$\text{WER} = \frac{\text{substitution} + \text{insertion} + \text{deletion}}{\text{reference}_{\text{length}}}. \tag{2}$$

由于莱文斯坦距离原始的设计是为了应用于计算机字符串代码的更正、而不是语言字词更正为目标，断然的应用莱文斯坦距离到 WER 来计算翻译质量自然产生误区。其中之一，词序因素在 WER 里没有得到合理的考虑，比如有的输出译文的词序和参考译文的词序相差比较大，在这种情况下 WER 为了计算字词的距离问题会给折扣分，但是因为语言的多样性以及词序的多变性，这个译文可以照常是一个好的译文。为了克服这一点，[147] 设计了自由词序的编辑距离 PER（Position-independent WER）。在 PER 的计算中，词的顺序被忽略掉，而只是匹配输出的词跟参考译文的词是否匹配。基于参考译文的长度和输出译文的长度对比，多余的输出词会被归为删词，而缺少的情况会被归为加词。PER 计算公式如下：

$$\text{PER} = 1 - \frac{\text{correct} - \max(0, \text{output}_{\text{length}} - \text{reference}_{\text{length}})}{\text{reference}_{\text{length}}}. \tag{3}$$

2006 年 [142] 的工作提出来另一种减少对词序不同的过度惩罚的解决方法，命名翻译编辑距离 TER（translation edit rate），这个方法更倾向于于回归莱文斯坦距离的原始设计，也就是–加、删、移动–，而不是 WER 里的–加、删、替换–。在这个设计里、如果输出译文的词序和参考译文不同，那么移动一个词（或者连续的词语块）从一个位置到另一个位置只计算一个惩罚分数而不是根据移动的距离来增减惩罚度。这个因素的设计也类似于模拟人工修改文件时候的剪切和粘贴步骤。TER 的设计也包含多个参考译文的情况，TER 计算如下（总编辑数除以参考译文平均长度）：

$$\text{TER} = \frac{\#\text{of edit}}{\#\text{of average reference words}} \tag{4}$$

*4.1.2 准确度和召回率*

准确度 (Precision) 和召回率 (Recall) 分别反应的是输出文本的准确率和对原文的忠实性，这两个标准是从信息检索的评价标准适应而来。

$$\text{Precision} = \frac{\#\text{of correct output}}{\#\text{of all output length}} \tag{5}$$

$$\text{Recall} = \frac{\#\text{of correct output}}{\#\text{of reference length}} \tag{6}$$

在早期流行的自动评测里，BLEU[123] 分数是基于 n 元语法的准确率来计算评价分数，BLEU 最初的设计是为了计算文档级别的相似性而非句子级别，因为翻译存在的稀疏性，四元以上的匹配在句子级别会导致分数大跌。常用

---
[5] 俄文首次出版的该文件发表于 1965 年，英文翻译第二年出版

的 BLEU 分数采用一元 (uni-gram) 到四元 (4-gram) 的译文和参考文的字词匹配，BLEU 还设置了一个简短惩罚系数（brevity penalty, BP）用以对倾向于输出短句子而提高准确率的机器翻译系统进行系数惩罚，计算如下：

$$BLEU = BP \times \exp \sum_{n=1}^{N} \lambda_n \log \text{Precision}_n \qquad (7)$$

$$BP = \begin{cases} 1 & \text{if } c > r, \\ e^{1-\frac{r}{c}} & \text{if } c <= r. \end{cases} \qquad (8)$$

其中 c 和 r 分别是输出译文 (candidate) 和参考译文 (reference) 长度。n 元准确度的权重系数 $\lambda_n$ 多设置为统一权重。2002 年为了改进 BLEU，增加对多元 n-gram 信息计算的权重，[42] 设计了 NIST 评测标准，在增加多元片段的计算权重的同时，还设计了算数平均数去代替 BLEU 的几何平均，并且扩展到 5 元字词计算（5-gram）。在 BLEU 的设计里面当有多个参考译文的情况下，与参考译文长度最近的译文会作为参考，而在 NIST 里平均译文的长度会被选择。NIST 信息权重计算如下：

$$\text{Info} = \log_2 \left( \frac{\#\text{occurrence of } w_1, \cdots, w_{n-1}}{\#\text{occurrence of } w_1, \cdots, w_n} \right) \qquad (9)$$

与基于准确度的 BLEU 和 NIST 相反，2005 年 [6] 提出来倾向于召回率的 METEOR 评测方法（召回率的权重设置是准确率的九倍）。在 METEOR 的字词匹配工程中，除了 BLEU 所用的精确匹配，基于词根 (stem) 和同义词 (synonym) 词表的匹配也算做有效匹配。在 METEOR 公示里，惩罚系数的计算设计是根据匹配字符串片段 (chunk) 的数量，如下：

$$\text{Penalty} = 0.5 \times \left( \frac{\#\text{chunks}}{\#\text{matched unigrams}} \right)^3, \qquad (10)$$

$$\text{MEREOR} = \frac{10PR}{R+9P} \times (1 - \text{Penalty}). \qquad (11)$$

更简洁地，[148] 在 2006 年设计完全的只依赖准确度和召回率的 F 方法指标 (F-measure)。F-measure 的计算是两者的调谐平均数 (harmonic mean) 如下：

$$F_\beta = (1 + \beta^2) \frac{PR}{R + \beta^2 P} \qquad (12)$$

当 $\beta$ 的取值为 1 的时候，两个因子的权重相同。

其他相关的评测模型包括 ROUGE[102]，是一个基于召回率的模型，最初被应用于文本总结 (summaries) 的评价，之后也被用作机器翻译评测 [103]。

*4.1.3 加强模型*

本节我们介绍对词序因子的重顾考虑，以及一些综合各类计算因子的加强的模型，包括我们自己提出的系列工作 [66, 74, 71]。在分别介绍了考虑编辑距离、准确度以及召回率之后，也有研究人员考虑将所有以上因素考虑进一个加强的评测标准里。基于这种考量，[155] 在 2009 年设计了 ATEC (assessment of text essential characteristics) 方法，在 ATEC 方法里，除了对准确度和召回率的设置，作者还引进了一个新的计算词序惩罚的因子 PDP（position difference penalty）。在字词的匹配过程中考虑了字面词、词根、发声、和词感 (sense) 的不同方面。相似地，结合准确率、召回率、和词序信息，[30] 开发了用于调整机器翻译系统参数的 PORT 标准，PORT 的设计是为了调整机器翻译使得翻译输出更准确。

基于以上工作的分析，于 2012 年，我们设计开发了加强模型 LEPOR[66]。LEPOR 的评测因素包含了准确度、召回率、改进的长度惩罚系数、和改进的词序惩罚系数 (起名于 "**L**ength Penalty, **P**recision, n-gram **Po**sition difference Penalty and **R**ecall")，计算如下：

$$\text{LEPOR} = LP \times NPosPenal \times Harmonic(\alpha R, \beta P) \quad (13)$$

$$LP = \begin{cases} e^{1-\frac{Length_{ref}}{Length_{hyp}}} & if\ Length_{hyp} < Length_{ref} \\ 1 & if\ Length_{hyp} = Length_{ref} \\ e^{1-\frac{Length_{hyp}}{Length_{ref}}} & if\ Length_{hyp} > Length_{ref} \end{cases} \quad (14)$$

$$NPosPenal = e^{-NPD} \quad (15)$$

$$NPD = \frac{1}{Length_{hyp}} \sum_{i=1}^{Length_{hyp}} |PD_i| \quad (16)$$

$$|PD_i| = |MatchN_{hyp} - MatchN_{ref}| \quad (17)$$

其中，*P* 和 *R* 分别代表准确度和召回率。*LP* 是改进的长度惩罚系数，*NPosPenal* 是基于 n 元对齐的位移（词序）惩罚因子（n-gram position difference penalty）。计算实例详见 [66, 74, 75]。

在最初的 LEPOR 设计上三个主要因子 (LP, NPosPenal, HPR) 的计算拥有相同的权重。考虑到不同语言在词序的要求上严格度不同，以及词语变化的多样性，我们认为应该对不同的因子分别设计权重参数，这样可以在实验开发集（development set）上对针对应用语言进行优化调参，以使得评估取得最佳结果。在这个思想指导下，在后续的更新工作里，我们还设计了完全调参的 hLEPOR，计算如下：

$$\text{hLEPOR} = Harmonic(w_{LP}LP, w_{NPosPenal}NPosPenal, w_{HPR}HPR) \quad (18)$$

$$HPR = \frac{(\alpha + \beta)Precision x Recall}{\alpha Precision + \beta Recall} \quad (19)$$

同时地，除了在词序惩罚的因子上考虑 n-gram 对齐，我们还设计了基于 n-gram 的准确度和召回率，这样我们得到 n-gram 的 LEPOR 起名 nLEPOR：

$$nLEPOR = LP \times NPosPenal \times exp(\sum_{n=1}^{N} w_n log HPR) \quad (20)$$

在 WMT2011 年的数据集上的测试，基于八个语言翻译方向的评测，包括英语 ↔ 捷克语/德语/西班牙语/法语，加强模型 LEPOR 击败 METEOR-1.3、BLEU、TER 和其他方法（AMBER, MP4IBM1）取得与系统级别人工评价的最好相关系数 0.77，其中包括捷克语到英语和西班牙语到英语的单语最高分数 0.95 以及 0.96 [66][6]

在 WMT2013 年的国际评测比赛上，在西班牙语到英语以及所有测试语言对的平均上，nLEPOR 取得片段级别（segment-level）评测方法的最高表现之一 (并列于 METEOR 和 sentBLEU-MOSES)，超越于其他团队提交的系统 [58]。在后续的语言学特征介绍里，我们会继续讲解结合句法信息的增强 hLEPOR 以及它在 WMT 的评测比赛表现。 hLEPOR 在 WMT2013 和 WMT2021 年 (TED 领域文本) 评测比赛中，分别取得英文到俄文系统级别评测的第一表现和最优等级表现 [69, 10, 81][7]；在 WMT2019 年的比赛中取得德语到捷克语翻译评测的最优等级表现，和人工评价的皮尔逊 (Pearson) 相关系数 0.959[8][8]。

4.2 基于语言学特征

本节我们介绍使用语言学信息来增强自动评价质量的方法，这包括句法和语义两个子类。在之前我们讲述的部分评测方法里面，比如 METEOR 和 ATEC 也有提到语言学特征（同义词、词根匹配），我们自行设计的 hLEPOR 方法也会应用到语言学信息。本节的设计是为了更系统的理解语言学信息在翻译评测上的使用。

4.2.1 句法信息

在句法 (syntax) 的信息里我们介绍词性 (part-of-speech, POS)、短语 (phrase)、和句子结构的成分在评测里的应用。这些信息的使用一般伴随着一些开源的词性标注和句法解析工具（比如 tagger, parser, 和 chunker），如图4。我们如下分别介绍。

在语法领域，词性（POS）"指单个词在一定的词类系统中的类别归属。一个词的词性是由一定的词类系统和该词自身的语法特性两方面决定的。" 常见的词性包括名词、动词、形容词、副词、介词、连接词、标点符号等。词

---

[6] https://github.com/lHan87/LEPOR
[7] 2021 年数据 https://github.com/poethan/cushLEPOR
[8] 数据下载 https://github.com/poethan/LEPOR

性信息在评测质量里的应用包括，2011 年 [127] 提出的基于源语言文本和机器翻译输出文本词性相似度和词素概率的模型 MP4IBM1，该模型的理论是基于 1993 年 IBM 公司提出的统计机器翻译五个模型里的第一个模型 (IBM Model-I)[18]。同时，另一研究组 [40] 开发了 TESLA 评测方法，结合双语词表同义词和词性的匹配信息。

区别于以上工作，我们自己的早期工作，基于 LEPOR 主要因子的协调平均而设计的 hLEPOR 算法，在只应用输出译文和参考译文（而非源语言文本）的词性序列相似度上进行计算 hLEPOR(POS)，其评价结果跟人工的相近程度在 WMW2011 年的数据集合上超过了 BLE、METEOR、MPF、AMBER、和 MP4IBM1。其测试语言对包括英语-法语、英语-德语（双向）。

之后我们结合字符串级别的 hLEPOR(word) 和词性级别的 hLEPOR(POS) 两类计算分数，从而得到混合加强的 hLEPOR(hybrid) 评价方法，该方法应用于自动输出译文和参考译文的字符串级别和词性标注级别上，在提交给 WMT2013 年的国际翻译评测比赛中，在所有提交的评测方法里（一共 18 个系统），在以英语为源语言的（English-to-other）系统评测上，包含英语-法语/德语/西班牙语/捷克语/俄语，与人工评测的相似度排名第一 [69, 71]，其基于 Pearson 的五个语言对平均相关系数准确度 0.85[12, 74]。这进一步反应了我们自行设计的 LEPOR 和 hLEPOR 评测方法所包括的可调权重的加强因子的有效性。hLEPOR 的开源工具在线可用[9]。

词组或短语，包括习语 (expression) 和组合词 (a group of words)、固定短语和非固定短语等，在语言学领域，是整个句子的重要组成部分。2011 年 [4] 的工作假设源语言文本和翻译输出的目标语言文本应该拥有相似的语法结构，其设计包含未知词、短语类型（包括名词短语、动词短语、和介词短语）和句子长度比例等，并专注于德文到英文的翻译评价。

2013 年，我们自己提出的基于源语言和翻译文本的短语类型匹配的评测 HPPR [73] 解释图见4和5。首先，我们对源语言好目标语言进行句子解析，然后分别提取出其短语序列（位于解析树里词性标注的上一层）。由于不同语种的解析标注之间有差异，所以短语序列集合会有不同解释，因此为了统一不同语言之间的解析标注集合，我们另外开展了跨语种短语标注的对齐工作 (phrase mapping)，提出了统一短语类型集 (universal phrase tagset)[70]。这样我们可以对法语和英语的短语标签进行到统一短语标签的对齐，然后我们方可在同一维度计算这两个句子的短语序列的相似度。计算相似度的算法为 HPPR，见公式：

$$HPPR = Harmonic(w_{Ps}N_1 PosPenal, w_{Pr}N_2 Pre, w_{Rc}N_3 Rec) \qquad (21)$$

其调和平均数的三个主因子在之前段落已有介绍，分别是基于 n 元模型的词序惩罚因子、准确度、和召回率。HPPR 不需要参考文本的参与，在 WMT2011 和 2012 年的法语到英语的数据集上实现系统级别 0.63 和 0.59 的与人工评价的相关性，接近于依赖参考译文的 BLEU 和 TER 的表现，但是因为 HPPR 不需要参考译文所以可以减少花费并应用于参考译文不可知的情况[10]。值得注意的是，我们设计的统一短语类型对齐拓展到了 21 种语言覆盖

---

[9] https://pypi.org/project/hLepor/

[10] HPPR 数据网站https://github.com/aaronlifenghan/aaron-project-hppr

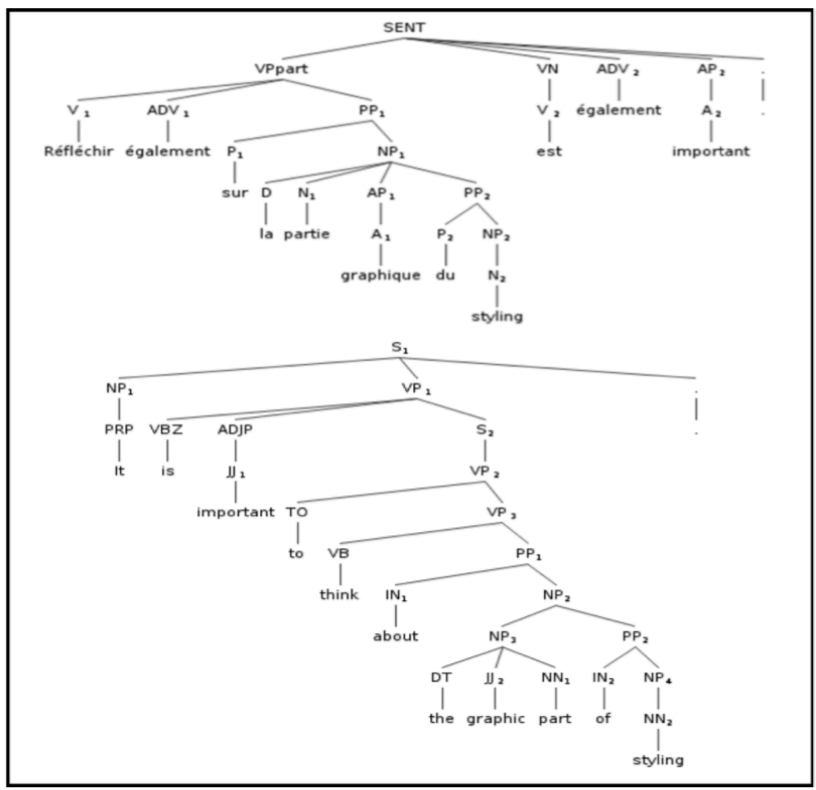

**Fig. 4** 英法两个平行句子的解析树 [73]

**Fig. 5** 英法两个平行句子短语序列到统一短语类型的对齐 [73]

25 个句子解析树库 (treebank)，因此不仅应用于英法，还包括中文、阿拉伯语、丹麦语、德语、意大利、日语、韩语、西班牙语等，详见 [70][11]。

句法在语法学中研究 "词的组合、句子结构和句子类型"。在介绍了词性和词语信息在评测里的应用之后，我们继续句子结构在此领域的使用。

---
[11] 统一短语类型对 25 个句法解析树库的标签对齐数据网站 https://github.com/aaronlifenghan/A-Universal-Phrase-Tagset

早在 1998 年，[131] 的研究工作探索机器翻译系统在翻译新文本类型时的表现，其专注于分析英语到丹麦语的翻译，并考虑句法上的信息，其中包括对介词短语和前置状语从句的识别。之后于 2005 年，[104] 的研究在考虑翻译译文的句子结构是否合理上，使用依存语法解析树 (dependency parser) 里的成分依赖信息，来计算依存解析树的相似性。此工作的实验证明使用句法信息有助于估计自动翻译译文的流利度。

其他有关应用句法信息的翻译评测扩展阅读包括应用词性标注的 [53, 130]，应用名词和动词短语类型的 [101, 45]，以及应用浅解析（shallow parser）的评测 [105] 和依存解析树的 RED[156] 模型。

### 4.2.2 语义信息

与侧重于词性、词组、和句子结构的句法信息相对应，语义学研究 "语言所表示的意义，包括词汇意义、语法意义和语用意义"。在本节我们介绍语义学特征在翻译评测里的使用。跨语言的语义计算可以直接应用于源语言文本和机器翻译输出目标语言文本的相似度，单语的语义计算则适用于测试机器翻译译文和参考译文的相似度。

首先，我们介绍命名实体（或命名体, named entity）在评测中的应用。命名体的使用引进于命名体识别领域，在命名体识别的自然语言处理任务上，通常定义的类型包括人名、地名、机构组织名、和时间实体等，参见 [113, 61] 以及我们自己的工作 [68, 76] 对此相关信息的介绍。2003 年，[5] 的评测实验工作佐证命名体特征的使用有利于机器翻译中改进句法分割以及消除歧义（消歧）的功效。在 2011 年的 WMT 评测比赛中，[90] 的工作分析，其基于统计机器翻译 Moses[90] 的模型，在命名体的检测上的失误导致系统性的流利度和准确度大跌。相似阶段，与 2012 年 WMT 的质量估计（QE）比赛上，[19] 分析，在其使用的所有特征集合里命名体对其评测结果向人工评测靠拢的贡献最大，其他特征包括编辑距离等。

其次，我们介绍组合词以及习语表达 (MWE, multi-word expression) 在翻译测试的使用。MWE 的研究有很多子领域包括合成词、习语、言语、成语、固定搭配等。也由于 MWE 的这些属性，与本世纪出开始，自然语言处理 (NLP) 学者指出 MWE 是多个任务领域的瓶颈所在 [135, 37, 78, 77, 79]。其中我们自己的非常近期的研究工作反应，相比较几个最前沿的神经翻译系统包括谷歌、微软、百度、和 DeepL，MWE 是目前翻译系统的一大难题，在富含 MWE 的句子上的自动翻译质量很差，尤其是针对 MWE 里的比喻、习语、外来词的翻译 [78, 77, 79]。

过去，在评测工作里使用 MWE 特征的研究还不多。于 2015 年，[136] 的工作主要针对 MWE 里的名词合成词以及组合词 (noun compounds, noun compound phrases) 进行研究，此工作使用斯坦福大学开放的句子解析器对名词合成词进行识别，并集成到输出译文和参考译文的相似性计算里。我们自己正在研究的评测模型 HiLMeMe(*Chapt*.5 in [72])（起名于 Human-in-the-loop MT evaluation looking into MWEs）将评测步骤分为两个阶段，其中包含句子级别的流利度与忠实度，以及短语级别的针对 MWE 的翻译质量，然后综合两个不同级别的分数生成最终的翻译评价。并且，针对 MWE 的挑战性以及现有公开测试文件对 MWE 涵盖的缺乏性（尤其是文学领域），我们开发了多语种

平行语料 AlphaMWE[77]¹², 此语料包括对 MWE 的标注，可用于 HiLMeMe 的测试应用。在本报告的未来研究发展部分，我们会再次提到对 MWE 的侧重学习。

再次，我们介绍同义词 (synonym) 的应用。同义词在自然语言处理领域可以是相同意思的词或者相近意思的词。广泛应用的同义词库包括 1990 年开放的 WordNet [119]。WordNet 的最初设计是一个专注于英文的大词库，把英语单词通过相近意思进行分组，主要针对四个词性名词、动词、形容词、和副词。同义词或同义词组会被归结到同一个 synset 单元里，每个 synset 单元里的词根据语义关系位于不同的层次等级排列。

然后，我们开始引入文本蕴含 (text entailment)。文本蕴含是指文字片段之间有方向性的蕴含关系。比如文本 A 成立的情况下，文本 B 会成立；但反之，不一定成立。与数理逻辑的严格推导不同，NLP 里的文本蕴含更是一种疏松的推导关系 [38, 39]。比如，在文本 A 成立的情况下，如果文本 B 大致成立或者在大部分情况下成立，则从文本 A 到 B 的蕴含关系一般可以建立起来。使用 WordNet 的语义特征, [29] 在 2012 年提出了一个基于文本蕴含的翻译评测方法，这个方法使用的是语义文本相似度 (semantic textual similarity, STS) 任务。

双向的文本蕴含会引出语义研究里的释义情况 (paraphrase)[3]。释义可以将一个句子或者一个段落里的字词做语义相同的动态处理 [115, 116, 11]。在 2006 年，[142] 所提出的 TERp (TER-Plus) 评测方法，采用了根据查询一个已存的释义表，来评价参考译文是否是机器输出译文的释义文本。

继续语义特征这一字节，我们介绍语义角色 (semantic roles) 的使用。研究者一般先对句子进行浅解析 (shallow parse)，和实体标注。然后用语义角色对参考译文和输出译文的观点和附属信息进行指代解释。比如 [53, 52] 的评测工作, 所使用的语义角色包括用使役代理 (causative agent)、副词辅助 (adverbial adjunct)、定向辅助 (directional adjunct)、否定标记 (negation marker)、和谓语辅助 (prediction adjunct)。相比较以上工作所用的一维（或平坦）语义角色 (flat semantic roles) 在一个进一步的工作里，[105] 提出结构性的语义角色关系 (structural relations) 比如谓词-参数关系 (predicate-argument relations) 新的 MEANT 评测方法，并且在 MEANT 里对不同的语义角色设置经验主义的不同权重。权重的经验设置来源于不同语义角色对保留和关联句子语义信息的重要程度。简言之，语义角色是句子和文本的语义结构的重要部分，并且在实验上被证明对评价翻译的忠实度有效。

在本节最后，我们引入语言模型 (language model)。统计语言模型一般包含词概率和次序概率的信息，从而可以对评测起到重要的知识存储信息支持。2005 年，[51] 的工作提出使用 "语言模型-支持向量机"(LM-SVM, support vector machine) 的方法在没有参考译文的情况下，估计译文输出的流利度。该工作目标在于训练一个分类器 (classifier)，从而区分出不流利和结构怪异的句子。当然该工作也属于自动评测里的质量估计的类型（QE）。随着近几年神经语言模型的推广，下一节我们会进一步介绍语言模型近期的应用。

总而言之，语言学信息（句法和语义）在评测里的使用包括两类，要么使用机器学习的方法把语言学知识作为训练特征来学习用以增强评测模型的知识量（如 [2, 98],），要么把句法和语义的特征直接集成到传统自动评测指标 (metric) 里来加强其测算量（如）[52, 143, 36, 74, 70, 75, 71]。

---

¹² github.com/poethan/AlphaMWE

4.3 基于深度学习

随着 2013 开始的词向量 (word vector, word to vector embedding) 和词组向量 (phrase embedding) 在 NLP 领域的流行和应用 [117, 118]，以及神经语言模型 (neural language model, NLM) 研究者的重回信心，基于深度神经网络 (DNN, deep neural network) 的神经机器翻译在几年之后（2015-2016 年）快速地跟上并超越了传统的统计机器翻译在大语料训练上的学习和表现水平 [107, 33, 83, 140, 149, 91]。与此同时，基于深度神经网络语言模型（DLM）的翻译评测也开始被研究人员所关注和使用。

早期应用神经网络的评测工作包括 [63]，此工作在神经网络的结构上使用简单的前馈神经网络 (feed-forward neural network, FFN)，在学习特征上使用了基于斯坦福神经解析器 (neural parser) 的句法信息和基于词向量 (Word2Vec, GloVe, COMPOSES 三个工具) 的句子级别语义信息，在机器学习模型的设计上使用成对比较的方法，也就分别计算两个不同机器翻译的输出与参考翻译的相近程度来选出较好的机器翻译，作者还探究了不同程度的前馈神经网络的深度对评测表现的影响，指出多层网络结果会更好，此工作的语料是针对以英语为目标语言的 WMT11-13 年数据。

相似的时期，在 2015 年，[62] 的评测工作 ReVal 提出使用树形 (tree-structure) 长短期记忆神经网络 (long short term memory, tree-LSTM) 来学习 WMT2013 年人工评测所赋予的不同机器翻译输出与参考翻译相比较的排序分数。在 WMT2014 年以英语为目标语言的测试集上表现良好。相似地，2016 年 [109] 的工作则使用了双向长短记忆神经网络 (Bi-directional LSTM) 进行特征学习；与 [63] 的成对比较选择不同，该工作在评测环节可以针对机器翻译输出和参考译文进行计算，不需要对不同翻译输出两两比较，减少计算复杂性。

伴随着基于注意力机制 (attention-based) 的神经语言模型的流行 [82, 149]，基于 DNN 的评测模型继续延伸，于 2019 年 [114] 的翻译评价工作在前人提出的 Bi-LSTM 的模型之上加入注意力机制（BiLSTM+Attention）用于学习输出译文和参考译文之间的关联。与此同时，作者还提出了借用自然语言推理 (natural language inference, NLI) 的模型 ESIM (Enhanced Sequential Inference Model [31]) 进行翻译评测，这个设置把自动输出翻译作为假设 (hypothesis) 而把参考译文作为结论 (premise)。此工作在 WMT2017 年的实验数据上取得优异表现，尤其是以英语作为目标语言的翻译语言对上（包括中文到英文），在以英语作为源语言的表现上则与其他先进的评测方法相比各有千秋。

随着基于多维度注意力机制 (encoder attentions, decoder attentions, and cross encoder-decoder attentions, 编码器、解码器、和链接此两者之间的注意力机制) 的 Transformer 在神经机器翻译的领军地位的诞生，双向 Transformer 结构的 BERT（Bidirectional Encoder Representations from Transformers）模型也由于其大规模预训练知识的应用 [41] 而在自然语言理解（natural language understanding, NLU) 任务上表现出非常好的效果。BERTscore 是 2020 年由 [157] 所提出的基于 BERT 以及与其相似的可以实现背景知识嵌入 (contextual embedding) 的预训练模型，对机器翻译输出的每个单位（token）进行基于背景文本的词嵌入同时对参考译文做相同处理，然后对机器翻译输出和参考译文的每个输出单位进行成对比较和 Cosine 相似度计算。考虑到每个单词在句子里的重要性指数可以不同，此工作还基于测试集文本使用 idf(inverse document frequency) 赋予给个单词以不同的权重因数[13]。最后的 BERTscore 是基于准确率、召回率、和调和平均 F 值。BERTscore 的训练集是 WMT2018 年 149 个机器翻译系统的输出、参考译文、和人工评价分数。

此后，2020 年 [139] 的基于 BERT 的工作提出 BLEURT 评测工具，此模型采用回译 (back-translation) 方法生成众多释义文本 (paraphrase) 以提高相似语义的词语表达覆盖度，并且集成 BLEU, ROUGE, BERTscore 三个评测指标作为其学习信号。在 WMT2017 年的测试数据上 BLEURT 的表现超过 BERTscore 和其他相似时期的基于深度学习的评测方法。

紧随其后，COMET 评测模型 [134] 提出使用多语种的预训练模型 (cross-lingual pre-trained LM, XLM)，以及在机器翻译输出和参考译文的两个因素外加入源语言文本作为学习因素。COMET 包含估计 (estimator) 模型和翻译排序 (translation ranking) 模型两个子块。其人工标注的训练数据是基于 HTER、MQM、和 DARR(direct assessment relative ranking) 三个类型。其在 WMT2019 年的测试集上在多个语言对上超过 BERTscore 和 BLEURT。

在不同的基于深度神经网络 (DNN) 和预训练模型（PLM）的新的评测方法的开展的同时，如何使用 DNN 和 PLM 对已有的传统评测方法优化是另辟蹊径的研究课题。比如 2016 年用词向量对 METEOR 进行加强的模型 [141]，以及我们自己提出的用 PLM 对 hLEPOR 的参数进行自动优化的模型 cushLEPOR (customising hLEPOR)。考虑到基于深度学习和预训练大数据的评测模型的成本花费大，我们提出的 cushLEPOR 工作使用蒸馏的技术 (distilled model) 用预训练模型对所应用的语言对进行 hLEPOR 评测模型的参数最优化，然后保留此参数对同样言语对的评测使用相同的参数集，以减少使用深度学习的评测成本以及保留传统字符匹配评测的快速便捷。cushLEPOR 的初步实验表明在针对 MQM 和预训练模型 LaBSE 的优化方法上，其在 WMT2021 年的评测比赛里取得新闻领域的最有表现评测方法之一 [47, 81]。其中，官方测试语言对包含英语-德语、中文-英语、以及英文-俄语[14]。

## 5 元评测 (评测的评测)

在元评测这一节，我们介绍针对评测方法的评测。这包括统计学重要性（也叫显著性差异, statistical significance），评价一致性 (agreement level)，和评测结果的相关系性数，以及对不同评测指标（metric）的相互比较等。

## 5.1 机器翻译显著性差异

当不同机器翻译系统在同一个数据集上表现不同的分数时，为了确信这样的不同是系统质量的真是反应，而不是随机性的，2004 年 [87] 的评测工作引入统计学里的显著性差异指标 (statistical significance)。这个工作使用自举重采样法（bootstrap re-sampling）来计算统计学显著性区间 (statistical significance intervals) 估计。这个方法在小样本的测试集上表现有效。显著性差异包含两个不同的策略，其一是概率值 p 值 (p-value)，此值计算给定一个 null 假设下

---

[13] idf 假设低频词 (rare words) 可能对翻译质量的反应比常用词 (common words) 更有效果。
[14] source code and parameters of cushLEPOR is available at https://github.com/poethan/cushLEPOR. WMT2021 shared task data: http://www.statmt.org/wmt21/

的观测数据出现的随机概率。另一个是类型-I（type-I）错误率，该错误率也称为假阳性错误（false positive），此指标计算错误地拒绝一个 null 假设来支持另一个假设的概率 [65]。

5.2 人工评价的可信度

人工评价一直以来被作为机器翻译质量的最终决定指标，人工评价的质量以及可信度在此领域起到至关重要的作用。为了减少人的主观偏见所导致的评测偏差，通用的设置是由两位以上人员评价相同的输出句子，如果不同人员对翻译输出的评价结果相似性（一致性，agreement）高，说明此人工评测结果比较客观公正。反之，如果不同人员的评价结果相差较大，那么这个评价结果作为标准评价的可信度会降低。除了不同人员之间的评价一致性水平 (inter-agreement)，同一评价者对句子质量的不同时间段的判断也可以计算自一致性 (intra-agreement)，比如，系统可以设置相同的机器翻译输出句子在不同时间段重复出现在评价者的任务列表里。对于评价一致性水平的计算方法，科恩的卡帕一致性系数（Cohen's kappa agreement coefficient）被 WMT 多年用作官方指标 [35]。其计算公式如下：

$$k = \frac{p_0 - p_c}{1 - p_c} \tag{22}$$

其中 $p_0$ 是评价者同意彼此的比率，$p_c$ 是评价者随机性同意彼此的比率 (expected by chance)。系数 $k$ 代表随机性的不一致不发生的比率；或者也可以说是除去随机性的一致以外的一致率。参数 $p_0 - p_c$ 代表除去了随机性的一致率 [92]。

从另一个方面来研究，被评价语料的大小有时候也会影响到对机器质量评价的客观性。上一个字节我们提到使用自举重采样法（bootstrap re-sampling）在小型语料情况下对机器翻译的输出进行扩充，最近我们自己提出的基于统计学采样法的模拟研究工作针对语料大小以及人工系统评测的偏差性进行蒙特卡洛统计模拟 (Monte Carlo Sampling Analysis)，用人工评价的编辑距离作为参数，估算人工评价设置下用以获得可靠评价的最小样本需求 [55]。此工作建议，在除去主观性评价的因素以外，准确的反应翻译机器翻译质量的语料不得小于 4000 字数。

5.3 自动评价与人工评价的相关性

对于自动评价模型的质量评价，一般是用其与人工评价的相关性系数高低来决定其优劣。本字节介绍 WMT 历年使用的相关性系数计算方法，不同的相关性系数的选择取决于输入参数是精确值还是排序值。

5.3.1 皮尔逊 (Pearson) 系数

皮尔逊相关性系数（Pearson's correlation coefficient）于 1990 年在 [126] 的工作中被介绍，通常用希腊字母 $\rho$ 来表示，用以计算两个随机变量 X 好 Y 的相关性，用 $\rho_{XY}$ 来表示 [120]，其计算方式如下：

$$\rho_{XY} = \frac{cov(X,Y)}{\sqrt{V(X)V(Y)}} = \frac{\sigma_{XY}}{\sigma_X \sigma_Y} \quad (23)$$

因为变量 $X$ 和 $Y$ 的标准差（standard deviations, $\sigma_X > 0, \sigma_Y > 0$）大于零，所以此两个变量的相关性指数（包括正、负、零相关）取决于它们的性方差 $\sigma_{XY}$ 的值。设定两个变量因子 $(X,Y)$ 的参数 $(x_i, y_i), i = 1\,to\,n$，皮尔逊系数可计算为：

$$\rho_{XY} = \frac{\sum_{i=1}^{n}(x_i - \mu_x)(y_i - \mu_y)}{\sqrt{\sum_{i=1}^{n}(x_i - \mu_x)^2}\sqrt{\sum_{i=1}^{n}(y_i - \mu_y)^2}} \quad (24)$$

其中 $\mu_x$ 和 $\mu_y$ 分别是两个变量的期望值。

斯皮尔曼 (Spearman) 等级相关性

斯皮尔曼的等级（排序）相关性是皮尔逊相关性系数的简化形式，并且被多年用于 WMT 的年度机器翻译评测报告 [23, 24, 25, 26, 69]。在没有平分（平局）的设定下，斯皮尔曼相关性 (rs) 可以计算为：

$$rs_{\varphi(XY)} = 1 - \frac{6\sum_{i=1}^{n} d_i^2}{n(n^2 - 1)} \quad (25)$$

其中 $d_i$ 是两个排序列的相应变量的差值（difference-value, $(x_i - y_i)$）。而变量序列 $\vec{X} = \{x_1, x_2, ..., x_n\}$ 和 $\vec{Y} = \{y_1, y_2, ..., y_n\}$ 则是对系统 $\varphi$ 的不同排序。

*5.3.2 肯德尔的 tau 一致性*

肯德尔的 tau 一致性(Kendall's $\tau$)[84] 在继斯皮尔曼相关性之后，被用于 WMT 的评测相关性计算，用于评测自动排序和参考排序的一致性 [25, 26, 27, 69]。其计算方式如下，分子为一致性的对数减去不一致的对数，分母为所有的成对比较数：

$$\tau = \frac{\text{num concordant pairs} - \text{num discordant pairs}}{\text{total pairs}} \quad (26)$$

肯德尔的 tau 的近代介绍包括 [85] 的工作，另外 [97] 于 2002 年综合回顾介绍了肯德尔的 $\tau$ 是如何被用于计算系统序列和参考序列的差异的。更精确地，[94] 在 2003 年引入 tau 在等级排序 (rank) 上计算系统生成的和人工提供的排序列的差异水平。

## 5.4 标准之间的互比较

不同评测标准之间的互比是另一个元评测的研究分支，比如早期的对 BLEU 评测标准的分析工作 [20, 22, 96] 指出，基于质化（qualitative）的研究表明，BLEU 并不能准确的翻译机器翻译质量的变化，较高的 BLEU 分数不能确保较好的翻译质量。相似的关于 BLEU 可信度的研究也反映在近期的研究工作，

比如来自谷歌研究团队的大规模人工后编辑分析 [50] 指出质量上比人工编辑后的翻译差的机器翻译输出可以会有高达 5 个点的 BLEU 值。因为 BLEU 对评测过程的语言学缺乏性，在机器翻译质量大大提升的平台下，这个乌龙现象不足为怪。

其他的代表性评测比较工作包括 [58, 56]，此研究团队在基于 WMT（2013, 2014）评测的历史数据上进行量化分析，指出近几年提出的加强的评测模型在句子级别（片段级别, segment-level）的挑战性任务上，比 BLEU 等传统流行评测标准要更好一个等级的工作包括 nLEPOR、SentBLEU-Moses 等。其中 nLEPOR 是我们自己研发的模型 [74, 71, 73]。

## 6 未来展望及研究方向

首先，组合词以及组合词表达 (multiword-expressions, MWEs) 的识别是自然语言处理的一个重要任务，组合词表达包含很多不同类别的词语组合并且涵盖比喻、言语、成语等成分，在机器翻译、自然语言处理 (NLP) 和评测任务扮演着非常重要的角色 [135, 110, 121, 132, 77]。这反映在历年的国际 MWE 研讨会和近几年该研讨会组织的 MWE 识别任务 [112, 111, 137]。因此，在此方面与机器翻译领域的两个未来研究方向包括：1) 组合词表达的识别模型和翻译模型的结合；2) 组合词表达在机器翻译评测里的应用。

针对 1)，目前在深度学习领域已有对 MWE 的可解析性 (decompositionality) 和可侦测性 (identification) 的前沿进展，比如 [64] 用神经网络对名词构成词的语义结构研究, 如何建立综合的神经网络模型 (hybrid neural networks) 将 MWE 构词和解析研究与神经机器翻译两个目前分开的神经学习模型结合起来，是一个非常可行的研究课题 (讨论见 [78])，并且这种结合的模型训练将有助于系统的总体优化，比如使用机器学习里现有的先进的神经网络高等参数优化框架模型 (hyper-parameter optimisation framework, Optuna)[1]。针对 2)，如何改进目前的广泛使用又饱受批评的流行评测方法（如 BLEU），设计新的评测模型，将语义评测合理囊括进评测系统里，这是一个很有挑战性的课题。而由于众多 MWE 子类对语义的涉及（比如言语、成语、习语），其在翻译评测过程扮演了一个很重要的角色。这既可以是障碍（如歧义）、也可以是助手。因此如何积极利用这一角色，发挥其优势是一个非常可行的研究方向，这包含相应的多语种对齐语料建设、人工标注、神经网络建模、和模型测试。

其次，篇章级别 (context-aware) 的机器翻译评测是未来发展趋势之一。目前的评测方法，大都关注于句子级别的内容。但是，从语言学角度观察，一个句子所在的环境（篇章背景）对本句子的理解起到至关重要的作用，尤其是含有指代词、和歧义词的情况，如何更精确的去翻译和评价翻译的好坏，需要对句子背后的信息有足够的认知掌握 [77]。这个研究方向在深度学习模型、以及相应的神经语言模型出现后变得非常可行。比如，目前成熟的深度学习模型可以不止对句子级别进行词到向量的转化，并且还可以对跨句子和篇章级别的内容进行向量转化 (sentences/paragraphs to vectors)，这样，对文本和背景知识的学习可以嵌入到评测系统里，作为模型学习的特征。

再次，基于具体任务 (task-oriented) 的翻译评测在机器翻译的大流行下变得越来越紧迫需要 [54]，比如旅馆预定的机器翻译，由于该领域句子偏短并且多附有表格填写，会更侧重于命名实体的翻译准确性如地名、机构名、人名（尤其外语人名的翻译）等；再比如目前刚开始流行的多模态机器翻译

(multi-modal MT）包括多模态图片标题生成和翻译（image captioning MT）任务，这样的情况下对多模态 (image+text) 资源的利用变得非常必要。

最后，无参考译文的机器翻译质量估计（QE）是研究的一个重点 [145, 67]。由于在某些情境下参考译文的缺失，比如地震灾害等情况下需要对当地语言进行多语种翻译以提高营救效果，无参考译文的质量估计模型更加的适用于现实的需求。这在 WMT 的历史机器翻译任务里有出现过。在其他情况，当参考译文的获得非常昂贵或者不实际时，没有参考译文的翻译质量信心估计也是一个挑战性的问题，比如现有的在线翻译平台软件，很少有在提供用户自动翻译译文的同时提供翻译质量估计水平 (confidence estimation)。在未来机器翻译和评测的发展中，如何将翻译和质量估计同时提供给使用者是一个难题。这涉及到翻译模型和质量估计模型的同步学习训练。

**7 结语**

此文主在介绍机器翻译评测的发展，内容覆盖人工评价模型、自动评价模型、元评测（评价模型的评价）、以及对此方向的未来发展研究展望。在人工评价和自动评价模型分块分别简要介绍了历史性的方法和前沿的进展，这包含人工评价里对 crowd-source 的应用以及自动评价里对当前的深度学习和预训练模型的运用。在元评测部分我们探讨了统计学中显著性差异、可信度等在评价里的应用、以及不同的相关性系数比较。由于机器翻译属于自然语言处理（NLP）的一个大的分支，涉及到自然语言理解 (NLU) 和自然语言生成 (NLG) 的其他不同子分支，我们希望这份综合性评测报告也会有利于其他 NLP 相关研究领域的推进、尤其在评测和质量估计建模方面，比如这包括摘要生成 (summarization) 的评测、图像标题生成 (image captioning) 的评测、释义 (paraphrase) 和蕴含 (entailment) 的评测、信息提取 (information extraction) 的评测、代码生成 (code generation) 的评测等。

**致谢**

26 Lifeng Han